\documentclass{article}
\usepackage{spconf,amsmath,graphicx}
\usepackage{amsmath}
\usepackage{amssymb}
\usepackage{bm}
\usepackage{bbding}
\usepackage[pagebackref,breaklinks,colorlinks]{hyperref}
\usepackage{cleveref}
\usepackage{xcolor}

\usepackage[letterpaper, top=1in, bottom=1.05in, left=0.62in, right=0.625in]{geometry}
\usepackage{tabularx} 
\usepackage{booktabs} 
\usepackage{graphicx} 
\usepackage{makecell}
\usepackage{makebox}
\usepackage{multirow}
\usepackage{comment}

\usepackage{enumitem}
\setlist{nosep, leftmargin=14pt}

\usepackage{mwe} 

\title{PGP-SAM: Prototype-Guided Prompt Learning for Efficient Few-Shot Medical Image Segmentation}
\name{Zhonghao Yan$^{1}$, \quad Zijin Yin$^{1}$, \quad Tianyu Lin$^{2}$, \quad Xiangzhu Zeng$^{3}$, \quad Kongming Liang$^{1 \ast}$\thanks{$\ast$ Corresponding author.}, \quad Zhanyu Ma$^{1}$}
\address{
    $^{1}$Beijing University of Posts and Telecommunications, Beijing, China \\
    $^{2}$Johns Hopkins University, Baltimore, USA
    \\
    $^{3}$Department of Radiology, Peking University Third Hospital, Beijing, China
}
\begin{document}

%
\maketitle
%
%
\begin{abstract}

The Segment Anything Model (SAM) has demonstrated strong and versatile segmentation capabilities, along with intuitive prompt-based interactions. However, customizing SAM for medical image segmentation requires massive amounts of pixel-level annotations and precise point- or box-based prompt designs. To address these challenges, we introduce PGP-SAM, a novel prototype-based few-shot tuning approach that uses limited samples to replace tedious manual prompts. Our key idea is to leverage inter- and intra-class prototypes to capture class-specific knowledge and relationships. We propose two main components: (1) a plug-and-play contextual modulation module that integrates multi-scale information, and (2) a class-guided cross-attention mechanism that fuses prototypes and features for automatic prompt generation. Experiments on a public multi-organ dataset and a private ventricle dataset demonstrate that PGP-SAM achieves superior mean Dice scores compared with existing prompt-free SAM variants, while using only 10\% of the 2D slices.

\end{abstract}

\section{Introduction}
\label{sec:intro}

Medical image segmentation aims to enable precise identification and delineation of specific regions of interest (e.g. tissues, organs), which can support doctors in making reliable diagnoses and planning effective treatments \cite{cheng2023sam,lin2024stable}. However, achieving accurate segmentation typically requires large volumes of annotated data, which demands significant time, expertise, and resources to collect and label.


\begin{figure}[htbp]
    \centering
    \includegraphics[width=\linewidth]{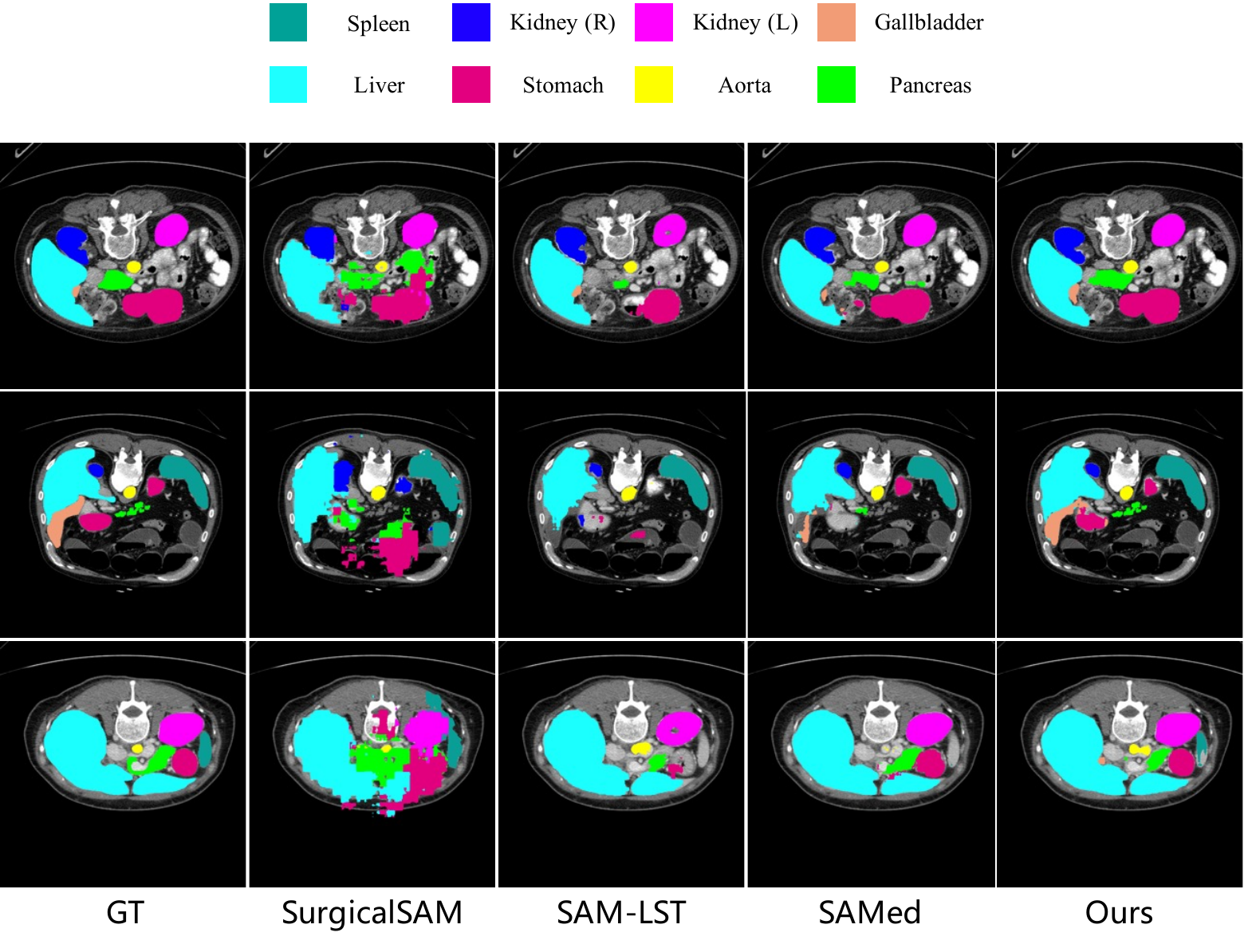}
    \vspace{-15pt}
    \caption{The qualitative results of PGP-SAM and other SAM variants, including prompt-free variant (SAM-LST and SAMed) and prompt-based variant (SurgicalSAM).}
    \vspace{-10pt}
    \label{fig:intro}
\end{figure}

The Segment Anything Model (SAM) \cite{kirillov2023segment} has brought significant advancements to image segmentation with its powerful segmentation abilities and user-friendly prompt-based interactions. However, its performance in zero-shot medical image segmentation has been reported as suboptimal in various studies \cite{he2023accuracy, mazurowski2023segment, cheng2023sam, wu2023self, ji2023sam}, spurring efforts to adapt SAM effectively to the medical domain.
Most current approaches focus on fine-tuning SAM’s image encoder to improve segmentation performance in medical imaging while preserving its interactive features. These methods often rely on ground truth data to create prompts during testing \cite{mazurowski2023segment, ma2024segment, feng2023cheap, cheng2023sam}. Additionally, prompt-free fine-tuning methods for SAM have emerged \cite{hu2023efficiently, chai2023ladder, zhang2023customized}, but they lack sufficient utilization of medical knowledge.. 
Despite significant advancements, these works encounter two main challenges: (1) Due to the considerable domain gap between natural and medical images, full or encoder-focused fine-tuning requires large volumes of annotated medical data. (2) Given SAM’s sensitivity to input point and box prompts, even slight prompt deviations can substantially affect results \cite{yue2023surgicalsam}, necessitating precise prompt designs during training and testing.
To address these challenges, H-SAM \cite{cheng2024unleashing} employs a two-stage hierarchical mask decoder while keeping the image encoder frozen. This strategy allows rapid integration of medical knowledge using only 10\% of available samples. However, due to the substantial number of learnable parameters, H-SAM still encounters difficulties in learning class-specific information effectively with such limited data. 

Following the previous few-shot setting, we introduce the \textbf{PGP-SAM}, a Prototype-Guided Prompt learning SAM, aiming at fast transfer class-specific and relations knowledge by prototype learning. PGP-SAM contains two sets of prototypes: intra-class prototypes and inter-class prototypes, which can learn class-specific representative knowledge and the shared knowledge of organs in CT images, respectively. They are updated during training through gradient back-propagation. PGP-SAM consists of two key modules: (1) \textbf{Contextual Feature Refinement}, which fuses contextual information along channel and spatial dimensions to focus features on regions of interest; (2) \textbf{Progressive Prototype Refinement}, which matches each intra-class prototype with the most similar inter-class prototypes and interacts with image features and class features to obtain enhanced prototypes for generating accurate prompts. By using prototypes, we can learn core features from a small amount of data. Our method achieves outstanding results on both a public dataset and an private dataset while introducing as few additional parameters as possible. 

In summary, our main contributions are:  
\begin{itemize}
  \item We propose an efficient contextual fusion mechanism that aids the model in better learning global information.  
  \item We design a prototype-based prompt encoder that provides accurate prompts for SAM without requiring any additional knowledge. 
  \item We achieve the state-of-the-art results on both a public dataset and a private dataset.
\end{itemize}

\begin{figure*}[htbp]
    \centering
        \includegraphics[width=0.95\linewidth]{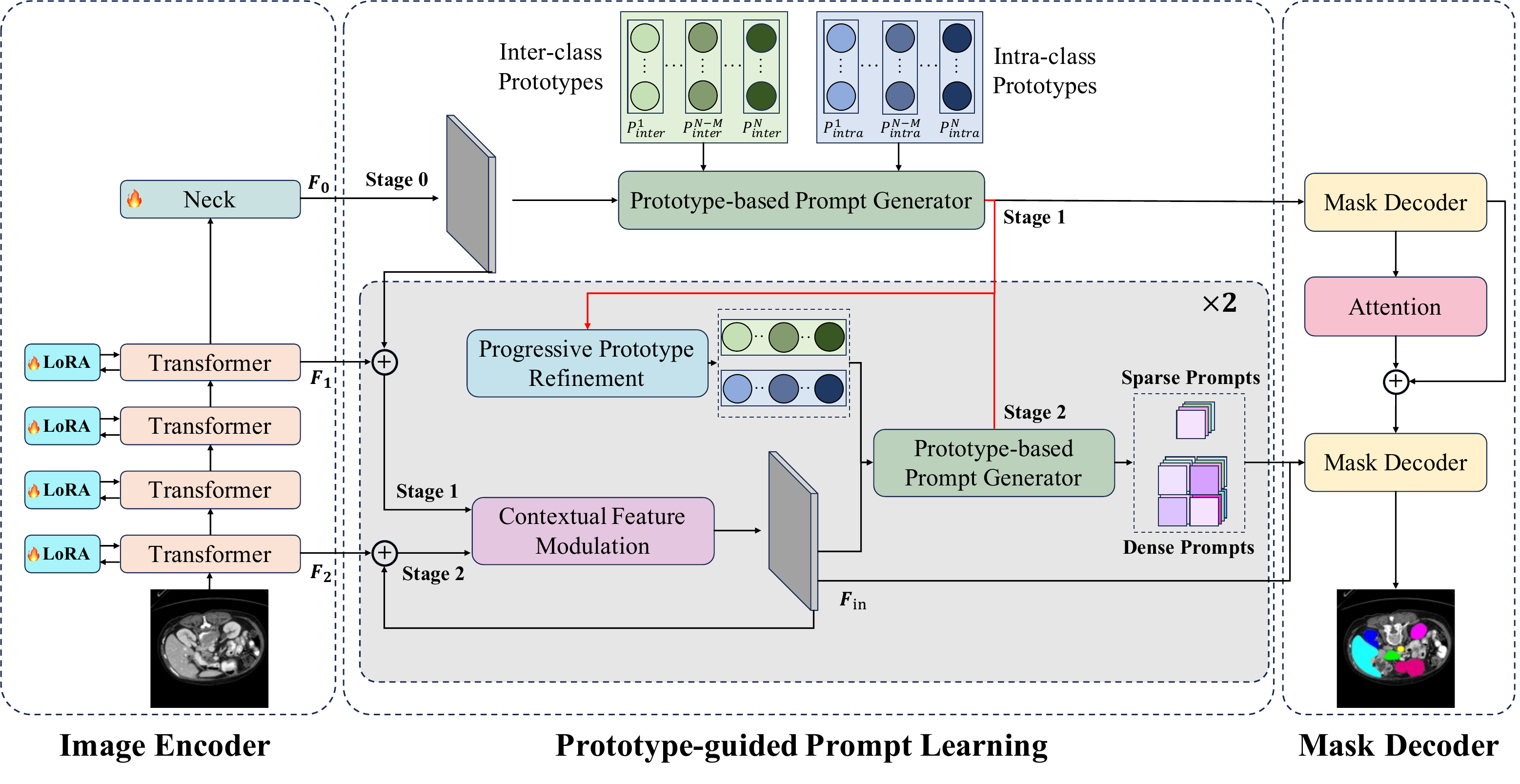}
    \vspace{-10pt}
    \caption{\textbf{Architecture of PGP-SAM.} As our main contribution, the Prototype-guided Prompt Learning Module consists of two identical stages and three key components: Contextual Feature Modulation (\textbf{CFM}), Progressive Prototype Refinement (\textbf{PPR}) and Prototype-based Prompt Generator (\textbf{PPG}).}
    \vspace{-12pt}
    \label{fig:model}
\end{figure*}

\section{Method}
\label{sec:method}


In this section, we introduce \textbf{PGP-SAM} by fine-tuning the image encoder of SAM. The proposed model comprises a multi-scale, prototype-guided prompt generator, and a mask decoder with shared weights that generates the segmentation mask. The details of PGP-SAM are illustrated in Figure~\ref{fig:model}.

\subsection{Contextual Feature Modulation}

The features extracted by the image encoder typically contain rich spatial details but often lack an overall image understanding. Inspired by previous work \cite{cavagnero2024pem}, we propose a Contextual Feature Modulation mechanism. This mechanism injects global semantic information into multi-scale features in both spatial and channel dimensions.

Given the input features $\bm{F}_i \in \mathbb{R}^{H \times W \times C}$ at each stage, with $i \in \{ 1, 2 \}$, we enhance them along both spatial and channel dimensions. We utilized features extracted after the first and fourth image blocks \cite{ke2024segment}. In the spatial dimension, $\bm{F}_s \in \mathbb{R}^{H \times W \times C}$ is obtained by applying average pooling operations along both the height and width directions and summing the results, which effectively enhances the features. Then, we compute the spatial feature modulation matrix $\bm{F}^s_i \in \mathbb{R}^{H \times W \times C}$:
\begin{equation}
  \bm{F}^{s}_i = \bm{F}_i \odot \delta (\varphi_{k \times 1}(\phi (\varphi_{1 \times k}(\bm{F}_s)))),
\end{equation}
where $\varphi$ indicates the strip convolution, and $k$ indicates the kernel size of the strip convolution. $\phi$ indicates the Layer Normalization followed by the ReLU function, and $\delta$ indicates the Sigmoid function. Similarly, we apply average pooling operations along the channel dimension to obtain $\bm{F}_c \in \mathbb{R}^{1 \times 1 \times C}$ and compute the channel feature modulation matrix $\bm{F}^c_i \in \mathbb{R}^{H \times W \times C}$:
\begin{equation}
  \bm{F^{c}_i} = \bm{F}_i \odot \delta (\varphi_{1 \times 1}(\phi (\varphi_{1 \times 1}(\bm{F}_c)))),
\end{equation}


Finally, by adding the spatially \(\bm{F}^s_i\) and channel-modulated \(\bm{F}^c_i\) features to \(\bm{F}_i\), we obtain the context-enhanced features \(\bm{F}_i\). For simplicity, subsequent calculations assume a single forward pass. Through global pooling along both the horizontal and vertical axes, CFM captures axial context, refines the network’s focus on foreground regions, and helps subsequent prototype learning identify the most class-relevant features. 

\subsection{Progressive Prototype Refinement}

The performance of SAM depends on accurate prompts, and even slight deviations can affect the segmentation results \cite{cheng2023sam, yue2023surgicalsam}. To address this, we introduce two sets of prototypes: intra-class prototypes $\bm{P}_{\text{intra}} \in \mathbb{R}^{N \times C}$, $\bm{N}$ is the number of class, and inter-class prototypes $\bm{P}_{\text{inter}} \in \mathbb{R}^{Q \times C}$, $\bm{Q} = \alpha \times \bm{N}$, $\alpha$ is set to $8$. Intra-class prototypes are used to learn unique information for each class, while inter-class prototypes capture information shared across all classes.

By computing the cosine similarity between \(\bm{P}_{\text{intra}}\) and \(\bm{P}_{\text{inter}}\), we first select the top \(k\) prototypes from \(\bm{P}_{\text{inter}}\) that are most similar to each class prototype in \(\bm{P}_{\text{intra}}\). These selected prototypes are then concatenated with \(\bm{P}_{\text{intra}}\) to form \(\hat{\bm{P}} \in \mathbb{R}^{Q \times C}\), where \(Q = N \times (k+1)\). 

Next, we design a class-based dual-path cross-attention mechanism to help \(\hat{\bm{P}}\) learn more robust class information. Specifically, given the enhanced input feature \(\bm{F}_I \in \mathbb{R}^{H \times W \times C}\) and the class feature \(\bm{F}_M \in \mathbb{R}^{H \times W \times N}\) from the previous stage, we calculate two sets of attention weights. The first set is the class attention weights \(\bm{W}_c\). We pass both \(\bm{F}_I\) and \(\bm{F}_M\) through pointwise convolutions to obtain \(\bm{F}_I' \in \mathbb{R}^{H \times W \times 1}\) and \(\bm{F}_M' \in \mathbb{R}^{H \times W \times 1}\). After reshaping, each is fed into a fully connected layer to produce \(\bm{F}_I'' \in \mathbb{R}^{H \times 1}\) and \(\bm{F}_M'' \in \mathbb{R}^{W \times 1}\).
\begin{equation}
  \bm{F}''_{\Theta} = \zeta(reshape(\varphi(\bm{F}_{\Theta}))), \quad \Theta \in \{ I, M \}
\end{equation}
These two outputs are then multiplied, followed by a pointwise convolution and a softmax operation, yielding the class attention weights \(\bm{W}_c \in \mathbb{R}^{H \times W \times Q}\). The process is formally described as:
\begin{equation}
  \bm{W}_{c} = softmax(\varphi(\bm{F}''_I \otimes \bm{F}''_M)),
\end{equation}
where $\varphi$ indicates the pointwise convolution, and $\zeta$ indicates the fully connected layer.


Next, $\bm{F}_I$ and $\bm{F}_M$ are fused through a pointwise convolution to obtain $\hat{\bm{F}} \in \mathbb{R}^{H \times W \times C}$. $\hat{\bm{P}}$ is used as the query, and matrix multiplication with $\hat{\bm{F}}$ is performed to obtain the query attention weights $\bm{W}_q \in \mathbb{R}^{H \times W \times Q}$. Finally, $\bm{W}_c$ and $\bm{W}_q$ are added together and multiplied with $\hat{\bm{F}}$ to obtain the enhanced $\hat{\bm{P}}' \in \mathbb{R}^{Q \times C}$ as:

\begin{equation}
\renewcommand{\arraystretch}{1.3}
    \begin{array}{cc} 
        \displaystyle \displaystyle \bm{W}_q = softmax(\hat{\bm{P}} \hat{\bm{F}}^T) \\
        \displaystyle \displaystyle \hat{\bm{P}}' = (\alpha \cdot \bm{W}_c + \beta \cdot \bm{W}_q) \odot \hat{\bm{F}}
    \end{array}
\end{equation}

We reshape $\hat{\bm{P}}'$ into $ N \times (k+1) \times C$, and then obtain $\hat{\bm{P}}_{\text{intra}} \in \mathbb{R}^{N \times C}$ and $\hat{\bm{P}}_{\text{inter}} \in \mathbb{R}^{N \times k \times C}$. $\hat{\bm{P}}_{\text{intra}}$ is added directly to $\bm{P}_{\text{intra}}$, while $\hat{\bm{P}}_{\text{inter}}$ is added to the corresponding $\bm{P}_{\text{inter}}'$ using the previously obtained top-k indices. 

By fully utilizing the image features and class features, the inter-prototypes and intra-prototypes we set can respectively learn class-specific information and inter-class information that is most similar to each class.

\subsection{Prototype-based Prompt Generator}

Previous work \cite{yue2023surgicalsam} computed similarity maps between prototypes and input features, then used these similarity maps to generate dense and sparse prompts. However, this approach led to the mixing of information. To address this, we designed separate pipelines for the dense and sparse prompts.



\textbf{Dense Prompt Generation.}
We first obtain \(\bm{P}_{\text{query}} \in \mathbb{R}^{Q \times N}\) by multiplying \(\bm{P}_{\text{inter}}\) and \(\bm{P}_{\text{intra}}\). Then, we concatenate \(\bm{P}_{\text{inter}}\) and \(\bm{P}_{\text{query}}\), pass them through a fully connected layer to get \(\hat{\bm{P}}_{\text{inter}} \in \mathbb{R}^{Q \times C}\). Next, we compute attention scores between \(\bm{F}_I \in \mathbb{R}^{H \times W \times C}\) and \(\hat{\bm{P}}_{\text{inter}}\), and feed the result into a two-layer MLP to form the dense prompt \(\bm{D} \in \mathbb{R}^{H \times W \times C}\):
\begin{equation}
  \bm{D} = MLP(softmax(\bm{F}_I \cdot \hat{\bm{P}}^T_{\text{inter}}) \cdot \hat{\bm{P}}_{\text{inter}}),
\end{equation}
where the $MLP$ consists of two pointwise convolutions, with a GELU activation function.

\textbf{Sparse Prompt Generation.}
Similarly, we rely on \(\bm{P}_{\text{intra}}\) to generate the sparse prompt. By concatenating \(\hat{\bm{P}}_{\text{intra}} \in \mathbb{R}^{N \times C}\) with \(\bm{F}_I\), we obtain:
\begin{equation}
  \bm{S} = MLP(softmax(\hat{\bm{P}}^T_{\text{intra}} \cdot \bm{F}_I) \cdot \bm{F}_I),
\end{equation}

By separating the pipelines for dense and sparse prompt generation, each prompt can better incorporate the knowledge learned from the prototypes, while reducing the risk of information overlap.

\begin{table*}[htbp]
\centering
\resizebox{1\linewidth}{!}{
  \centering
  \belowrulesep=0pt
  \aboverulesep=0pt
  \begin{tabular}{l|cccccccc|c}
    \toprule
     Method & Spleen & \makecell*[c]{Right\\Kidney} & \makecell*[c]{Left\\Kidney} & Gallbladder & Liver & Stomach & Aorta & Pancreas & \makecell{Dice$\uparrow$ (\%)} \\
     \midrule
     FSSP-SAM {\fontsize{8}{14}\selectfont \textcolor[gray]{0.8}{MICCAI2024Workshop}} \cite{wu2023self} & 72.84 & 44.93 & 48.00 & 19.86 & 72.76 & 29.86 & 14.55 & 11.36 & 38.90 \\
     AutoSAM {\fontsize{8}{14}\selectfont \textcolor[gray]{0.8}{Arxiv2023}} \cite{hu2023efficiently} & 68.80 & 77.44 & 76.53 & 24.87 & 88.06 & 52.70 & 75.19 & 34.58 & 55.69  \\
     SurgicalSAM {\fontsize{8}{14}\selectfont \textcolor[gray]{0.8}{AAAI2024}} \cite{yue2023surgicalsam} & 57.31 & 57.49 & 64.10 & 20.65 & 64.64 & 47.44 & 71.04 & 23.48 & 59.87  \\
     SAM-LST {\fontsize{8}{14}\selectfont \textcolor[gray]{0.8}{Arxiv2023}} \cite{chai2023ladder} & 79.91 & 81.26 & 82.04 & 35.05 & 92.35 & 52.15 & 73.14 & 29.44 & 65.67 \\
     SAMed {\fontsize{8}{14}\selectfont \textcolor[gray]{0.8}{Arxiv2023}} \cite{zhang2023customized} & 87.06 & 86.23 & 89.42 & 34.43 & 93.80 & 77.06 & 83.02 & 42.87 & 73.91  \\
     PGP-SAM (ours) & \textbf{89.49} & \textbf{90.11} & \textbf{91.45} & \textbf{49.14} & \textbf{95.23} & \textbf{78.77} & \textbf{86.83} & \textbf{54.16} & \textbf{78.75} \\
    \bottomrule
  \end{tabular}
  }
  \vspace{-10pt}
  \caption{Quantitative results on Synapse multi-organ CT dataset with few-shot settings.}
  \vspace{-8pt}
  \label{tab:synapse}
\end{table*}

\begin{table*}[htbp]
\centering
\begin{minipage}{0.7\linewidth}  
\centering
\resizebox{\linewidth}{!}{
  \belowrulesep=0pt
  \aboverulesep=0pt
  \begin{tabular}{l|cccc|c}
    \toprule
     \maketitle{Method} & \makecell*[c]{Right\\Ventricle} & \makecell*[c]{Left\\Ventricle} & \makecell*[c]{Third\\Ventricle} & \makecell*[c]{Fourth\\Ventricle} & \makecell{Dice$\uparrow$ (\%)} \\
     \midrule
     FSSP-SAM {\fontsize{8}{14}\selectfont \textcolor[gray]{0.8}{MICCAI2024Workshop}}  \cite{wu2023self} & 19.67 & 24.84 & 10.28 & 12.13 & 21.04 \\
     SurgicalSAM {\fontsize{8}{14}\selectfont \textcolor[gray]{0.8}{AAAI2024}} \cite{yue2023surgicalsam} & 32.53 & 35.41 & 15.69 & 27.29 & 30.73 \\
     AutoSAM {\fontsize{8}{14}\selectfont \textcolor[gray]{0.8}{Arxiv2023}} \cite{hu2023efficiently} & 59.69 & 59.18 & 40.33 & 45.36 & 52.38 \\
     SAM-LST {\fontsize{8}{14}\selectfont \textcolor[gray]{0.8}{Arxiv2023}} \cite{chai2023ladder} & 74.93 & 72.40 & 60.80 & 54.13 & 62.42 \\
     SAMed {\fontsize{8}{14}\selectfont \textcolor[gray]{0.8}{Arxiv2023}} \cite{zhang2023customized} & 80.87 & 77.35 & 64.50 & 70.65 & 73.36 \\
     PGP-SAM (ours) & \textbf{83.29} & \textbf{81.02} & \textbf{67.73} & \textbf{73.42} & \textbf{76.39} \\
    \bottomrule
  \end{tabular}
  }
  \vspace{-10pt}
  \caption{Quantitative results on Ventricle CT dataset with few-shot settings.}
  \vspace{-10pt}
  \label{tab:ventricle}
\end{minipage}%
\hfill
\begin{minipage}{0.28\linewidth}  
\centering
\resizebox{\linewidth}{!}{
  \belowrulesep=0pt
  \aboverulesep=0pt
  \begin{tabular}{ccc|cc}
    \toprule
     CFM & PPG & PPR & \makecell{Dice$\uparrow$ (\%)} \\
     \midrule
     \XSolidBrush & \XSolidBrush & \XSolidBrush & 71.92 \\
     \Checkmark & \XSolidBrush & \XSolidBrush & 74.08 \\
     \Checkmark & \Checkmark & \XSolidBrush & 75.32 \\
     \Checkmark & \Checkmark & \Checkmark & 78.75 \\
    \bottomrule
  \end{tabular}
  }
  \vspace{-10pt}
  \caption{Effectiveness of three key modules in PGP-SAM.}
  \vspace{-10pt}
  \label{tab:ablation}
\end{minipage}
\end{table*}

\section{Experiments}
\label{sec:experiments}

This section evaluates PGP-SAM on two segmentation tasks, comparing it with existing SAM variants and analyzing module contributions through an ablation study.

\subsection{Datasets}

We conducted multi-organ segmentation experiments on both the public Synapse multi-organ CT dataset \cite{landman2015miccai} and an private Ventricle CT dataset. The Synapse dataset contains 18 training cases and 12 testing cases, with each case having approximately 70 valid slices, while the Ventricle dataset has 400 training and 100 testing cases, each with around 10 valid slices. In the few-shot training, both datasets were at a resolution of \(512 \times 512\). 

\subsection{Implementation Details}

All models were trained on a single RTX 3090 GPU with data augmentation (elastic deformation, rotation, scaling). The loss function combines Cross-Entropy and Dice. We fine-tuned the ViT-b model using the same LoRA settings (rank=4) as SAMed. For the few-shot task, only 10\% of the training data was used, posing significant challenges without reference images. The optimizer is based on AdamW with \(\beta_1=0.9\), \(\beta_2=0.999\), and weight decay of 0.1.

\subsection{Results and Analysis}

In Table \ref{tab:synapse}, PGP-SAM shows outstanding few-shot transferability with limited medical images ($10\%$ of the total valid slices). Compared to other SAM medical SAM variants: FSSP-SAM \cite{wu2023self}, AutoSAM \cite{hu2023efficiently}, SurgicalSAM \cite{yue2023surgicalsam}, SAM-LST \cite{chai2023ladder} and SAMed \cite{zhang2023customized}, our PGP-SAM achieved a result of $78.75\%$ under limited training slices , outperforming other Medical SAM models. We further validated the effectiveness of our method on an private dataset in Table \ref{tab:ventricle}. The Ventricle CT dataset contains approximately 400 head scans from patients with varying degrees of brain hemorrhage and trauma. The presence of hemorrhage complicates the segmentation of ventricles, particularly in few-shot scenarios. Compared to other SAM prompt-free variants, our PGP-SAM achieved a result of $76.39\%$.
As shown in Figure \ref{fig:intro}, we visually compare the segmentation performance of PGP-SAM and other SAM variants on the Synapse dataset. Compared to the prompt-free variant, our method significantly reduces false negatives and positives by utilizing prototypes. Compared to existing sam variants, our approach provides more precise boundary delineation for segmented objects.

\subsection{Ablation Study}

To validate the effectiveness of our approach, we analyzed the three key modules of PGP-SAM, as shown in Table \ref{tab:ablation}. (1) \textbf{Contextual Feature Modulation (CFM)} improves the baseline by $2.16\%$. By leveraging early features from the Image Encoder, CFM enhances the model’s ability to capture fine details such as object boundaries and textures. (2) With the addition of the \textbf{Prototype-based Prompt Generator (PPG)}, the model’s performance increases by $1.24\%$, thanks to a tailored pipeline for generating dense and sparse prompts, designed to avoid information blending and ensure specificity. (3) The addition of \textbf{Progressive Prototype Refinement (PPR)} yields a further $3.43\%$ improvement. By allowing two prototypes to learn intra-class and inter-class information separately, PPR maximizes essential feature utilization from limited data. Even without reference images, this module greatly strengthens model robustness while maintaining high segmentation precision in data-scarce environments.

\section{Conclusion}
\label{sec:Conclusion}



We propose PGP-SAM, an efficient prototype-guided prompt encoder for adapting the Segment Anything Model to medical image segmentation. By generating accurate prompts without extra overhead, it addresses the model’s  under-utilization of medical knowledge and excels in few-shot tasks, highlighting its potential in data-scarce scenarios.

\section{Acknowledgments}
\label{sec:Acknowledgments}

This work was supported by the Beijing Natural Science Foundation Project No. L242025, the National Nature Science Foundation of China (Grant 62225601, 62476029, U23B2052), and sponsored by Beijing Nova Program.



\bibliographystyle{IEEEbib}
\bibliography{strings,refs}

\end{document}